\documentclass[letterpaper, 10 pt, journal, twoside]{ieeetran} 

\IEEEoverridecommandlockouts                              

\usepackage{graphicx} 
\usepackage{booktabs}
\usepackage{epsfig} 
\usepackage{amsmath} 
\usepackage{amssymb}  
\usepackage{tikz}
\usepackage{cite}
\usepackage{hyperref}
\usepackage{makecell}
\usepackage{leftindex}
\usepackage{subcaption}
\usepackage{svg}
\usepackage{bm}
\usepackage{xcolor}
\usepackage{siunitx}
\usepackage{pgf}   
\usetikzlibrary{shapes.geometric, positioning, fit, arrows.meta}
\usetikzlibrary{shapes.geometric, arrows.meta, positioning}

\newcommand\conf{\bm{q}}
\newcommand\state{\bm{x}}
\newcommand\control{\bm{u}}
\newcommand\velocity{\bm{v}}

\newcommand\goal{\bm{p_\text{goal}}}
\newcommand\eepos{\bm{p_\text{ee}}}

\newcommand\traj{\bm{Q}}
\newcommand\noisytraj{\bm{\hat{Q}}}
\newcommand\refinedtraj{\bm{\hat{Q}^\star}}

\newcommand\statetraj{\bm{X}}
\newcommand\noisystatetraj{\bm{\hat{X}}}

\newcommand\controltraj{\bm{U}}

\newcommand\x{\bm{x}} 

\title{\LARGE \bf
Warm-Starting Collision-Free Model Predictive Control With Object-Centric Diffusion
}

\author{Arthur Haffemayer$^{1,2,3, *}$,Alexandre Chapin$^{4}$,  Armand Jordana$^1$, Krzysztof Wojciechowski$^{1}$, Florent Lamiraux$^{1}$, Nicolas Mansard$^{1, 2}$, Vladimir Petrik$^{5}$ \vspace{-20pt}
\thanks{Manuscript received: September 26, 2025; Revised: December 12, 2025; Accepted: January 14, 2026.}
\thanks{This paper was recommended for publication by Editor Aniket Bera upon evaluation of the Associate Editor and Reviewers’comments.}
\thanks{This work is supported by the European project AGIMUS (under GA no.101070165), ANITI (ANR-19-P3IA-0004), and ELLIOT (No. 101214398)}
  \thanks{\textsuperscript{1}\, LAAS-CNRS, Universit{\'e} de Toulouse, CNRS, Toulouse, France}%
\thanks{\textsuperscript{2}\, Artificial and Natural Intelligence Toulouse Institute, France}%
\thanks{\textsuperscript{3}\, Continental, France}%
\thanks{\textsuperscript{4}~Ecole Centrale de Lyon, LIRIS, France}%
\thanks{\textsuperscript{5}~CIIRC, Czech Technical University in Prague}%
\thanks{\textsuperscript{*}{corresponding author}:
    \href{mailto:arthur.haffemayer@laas.fr}{arthur.haffemayer@laas.fr}}
\thanks{Digital Object Identifier (DOI): see top of this page.}

}

\begin{document}

\IEEEpubid{\begin{minipage}{\textwidth}\ \\[25pt]
\centering
© 2026 IEEE. Personal use of this material is permitted. Permission from IEEE must be obtained for all other uses, in any current or future media, including reprinting/republishing this material for advertising or promotional purposes, creating new collective works, for resale or redistribution to servers or lists, or reuse of any copyrighted component of this work in other works.
\end{minipage}}

\maketitle 

\begin{abstract}

Acting in cluttered environments requires predicting and avoiding collisions while still achieving precise control. Conventional optimization-based controllers can enforce physical constraints, but they struggle to produce feasible solutions quickly when many obstacles are present. Diffusion models can generate diverse trajectories around obstacles, yet prior approaches lacked a general and efficient way to condition them on scene structure. In this paper, we show that combining diffusion-based warm-starting conditioned with a latent object-centric representation of the scene and with a collision-aware model predictive controller (MPC) yields reliable and efficient motion generation under strict time limits. Our approach conditions a diffusion transformer on the system state, task, and surroundings, using an object-centric slot attention mechanism to provide a compact obstacle representation suitable for control. The sampled trajectories are refined by an optimal control problem that enforces rigid-body dynamics and signed-distance collision constraints, producing feasible motions in real time. On benchmark tasks, this hybrid method achieved markedly higher success rates and lower latency than sampling-based planners or either component alone. Real-robot experiments with a torque-controlled Panda confirm reliable and safe execution with MPC. An open-source implementation is provided \href{https://ahaffemayer.github.io/diffusion_warmstart_slot/}{here}.

\end{abstract}


\section{INTRODUCTION}

Generating collision free and dynamically feasible motions remains a central challenge in robotic manipulation, especially in cluttered and non convex environments. Classical sampling based planners explore high dimensional spaces effectively but do not handle dynamics \cite{lavalle_rapidly-exploring_1998, karaman_sampling-based_2011}, while trajectory optimization and MPC based methods explicitly incorporate smoothness, dynamics and collision constraints \cite{gaertner_collision-free_2021, haffemayer_collision_2025}. Despite recent progress in collision free MPC, these solvers remain highly sensitive to initialization and often converge to local minima in complex scenes.

Learning-based trajectory priors have recently emerged as a way to address this limitation.
In particular, diffusion models \cite{ho_denoising_2020} have shown strong ability to capture the multimodal structure of feasible trajectories, where several distinct solutions may exist around obstacles. When combined with trajectory optimization, these priors help guide the solver toward promising regions of the search space \cite{seo_presto_2025,huang_diffusion-based_2023}. However, to generalize to unseen environments, diffusion models must be conditioned on scene information \cite{huang_diffusionseeder_2024}, and designing an effective and compact representation of obstacles remains a key challenge.

\IEEEpubidadjcol

\begin{figure}[t]
    \centering
    \includegraphics[width=0.8\columnwidth]{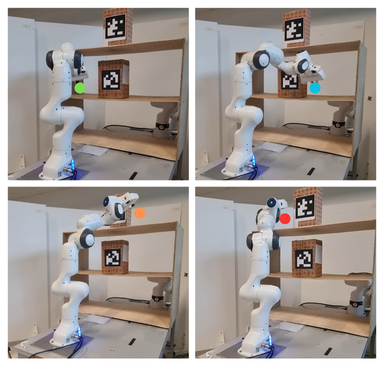}
    \caption{
    \textbf{Collision-free MPC with diffusion-generated warm-start.} The robot, sequentially reaches multiple target end-effector positions (colored dots) using collision-free MPC. The learned priors allow effective control within the non-convex space of the cluttered shelf environment.
    }
    \label{fig:teaser}
\end{figure}

\begin{figure*}
    \centering
    \includegraphics[width=0.85\linewidth]{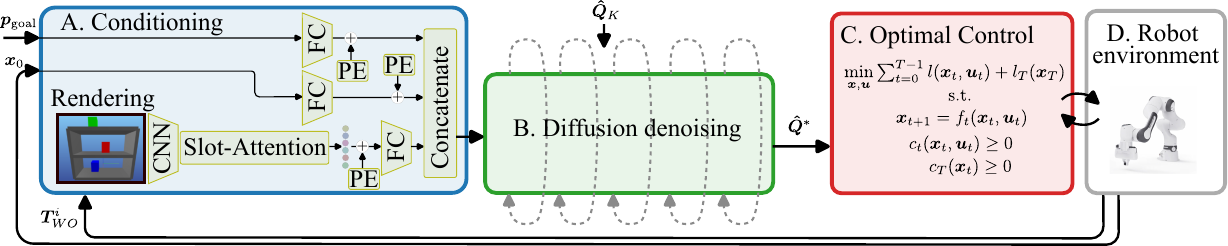}
    \caption{
     \textbf{Overview of the proposed architecture for robot control.}
    Our objective is to control a robot (D) using an MPC feedback loop that solves an optimal control problem (C) over the robot's dynamics, denoted by $f_t(\cdot)$, while strictly enforcing obstacle-avoidance constraints $c_t(\cdot,\cdot)$. Effective warm-starting is crucial for enabling the MPC solver to converge to a feasible, collision-free solution. To this end, we employ a diffusion model (B) that denoises an initial noise sequence $\noisytraj_K$ to generate a warm-start trajectory $\refinedtraj$, thereby accelerating convergence. To incorporate environmental context, we propose an object-centric conditioning mechanism (A) based on Slot Attention. This conditioning takes as input the target end-effector position~$\goal$, the current robot state~$\x_0$, and a latent representation of the environment derived from the estimated object poses $\bm T^i_{WO}$ via Slot Attention. The Slot Attention model internally renders a synthetic image from these poses, which helps mitigate the real-to-sim gap between simulated training data and real-world deployment.
    }
    \label{fig:overview}
\end{figure*}
From this perspective, we first argue that an object-centric representation of the scene provides a compact and highly informative conditioning signal for trajectory diffusion. We propose a Slot-Attention–based object-centric encoding that extracts a small set of latent obstacle descriptors from estimated object poses \cite{locatello_object-centric_2020}. This representation captures the essential collision structure while enabling fast inference and improved generalization compared to alternative conditioning strategies. We then show that the resulting scene-conditioned diffusion model interacts naturally with a constraint-based MPC solver, which rigorously and efficiently projects the diffused trajectory prior onto a dynamically feasible and collision-free trajectory. In contrast to previous approaches that rely on simple gradient-based post-processing of diffusion samples, our diffusion prior is refined through full OCP projection running in real time on a physical robot. In addition, we introduce a systematic dataset-generation procedure that combines sampling-based planning with constrained trajectory optimization to produce high-quality supervision trajectories.

By coupling these three components, object-centric scene representation, conditioned trajectory diffusion, and real-time MPC refinement, we demonstrate collision-aware manipulation in cluttered, non-convex environments. Taken together, these elements yield a complete robotics pipeline that demonstrates robust real-robot performance and achieves consistent improvements over existing baselines. The experimental analysis further illustrates how each component contributes to the overall capability of the approach.

\section{RELATED WORK}\label{sec:sota}

\subsection{Classical and learning-based motion planning methods}
Classical sampling-based planners \cite{reif_complexity_1979, canny_complexity_1988}, explore high-dimensional configuration spaces effectively but do not account for system dynamics \cite{lavalle_rapidly-exploring_1998, lamiraux_prehensile_2022}. Trajectory optimization methods \cite{ratliff_chomp_2009, kalakrishnan_stomp_2011, schulman_motion_2014} address this limitation by explicitly incorporating smoothness and dynamic constraints. MPC further extends this approach, formulating collision avoidance as either a hard constraint or a penalization in the optimal control problem \cite{mayne_model_2014, gaertner_collision-free_2021, haffemayer_collision_2025}, allowing for online planning of dynamically feasible, collision-free motions. Alternative approaches such as Model Predictive Path Integral control \cite{sundaralingam_curobo_2023, xue_full-order_2025} and tree-search methods like Monte Carlo Tree Search \cite{bonanni_monte_2025} introduce stochastic exploration to mitigate local minima, but their computational cost often limits real-time applicability. Despite these advances, classical TO, MPC, and sampling-based extensions remain sensitive to initialization and prone to local minima in complex, non-convex environments.

Learning-based methods provide a complementary solution by leveraging prior experience from demonstrations \cite{kober_imitation_2010, mandlekar_what_2021} or previously successful trajectories \cite{ichter_learned_2019, qureshi_motion_2019, urain_learning_2022, jordana_infinite-horizon_2025}, guiding planners or optimizers toward promising regions of the search space. In particular, providing a high-quality warm-start near a feasible solution can dramatically accelerate convergence and improve final solution quality \cite{stolle_policies_2006, mansard_using_2018, lembono_learning_2020}. Achieving such warm-starts is challenging due to the high degree of multi-modality in robotic trajectory generation, where multiple distinct solutions may exist for the same task.

\subsection{Diffusion models in robotics}
Diffusion models \cite{ho_denoising_2020} have recently emerged as a powerful tool to capture the multi-modal distributions of feasible trajectories. Early work such as Diffuser \cite{janner_planning_2022} iteratively denoises trajectories using a learned model and differentiable guiding functions, while Diffusion Motion Planning \cite{carvalho_motion_2023} adds gradient-based guidance for task-specific objectives, such as collision avoidance or smoothness. Extensions incorporating model-based projections \cite{romer_diffusion_2025} further improve the quality of generated trajectories. Composable Diffusion Models \cite{power_sampling_2023} learn distributions over constraint-satisfying trajectories and leverage them to handle novel combinations of constraints. While effective in low-dimensional or structured problems, these approaches generally struggle to generalize to cluttered or unstructured scenes. \cite{huang_toward_2024} recently applied diffusion models to approximate globally optimal nonlinear MPC. Their approach generates a diverse offline dataset of optimal control sequences, trains a diffusion model to capture the multi-modal distribution of solutions, and samples near-globally optimal trajectories at runtime without relying on initial guesses. While effective in simplified simulated environments, it has not been demonstrated on real robots. 

\subsection{Scene-conditioned diffusion approaches}
To improve generalization to unseen environments, several works condition diffusion models directly on scene information \cite{chisari_learning_2024, tian_warm-starting_2025, huang_diffusion-based_2023, seo_presto_2025}. Diffusion Seeder \cite{huang_diffusionseeder_2024} goes a step further by deploying diffusion-generated trajectories on a real robot; however, like prior work such as \cite{carvalho_motion_2023}, it generates only open-loop trajectories that are tracked by a low-level controller. These approaches do not perform closed-loop replanning, enforce dynamic feasibility during generation, or integrate with a receding horizon MPC framework. While they improve robustness compared to purely simulated methods, they remain limited to simplified or low-dimensional setups and do not demonstrate real-time, collision-aware control with compliance.

\subsection{Environment representation}
A key factor in scene-conditioned planning is how the environment is represented. Image-based representations \cite{chi_diffusion_2024} can lose geometric fidelity, whereas point clouds \cite{chisari_learning_2024, tian_warm-starting_2025} capture 3D geometry but are high-dimensional, unstructured, and require specialized networks to extract object-level features. Similarly, Diffusion Seeder \cite{huang_diffusionseeder_2024} relies on depth images, which provide direct 3D information but also suffer from high dimensionality and noise, limiting inference speed and generalization. Structured scene representations, such as PRESTO \cite{seo_presto_2025}, encode key robot configurations labeled for collision status, while object-centric approaches like Slot-Attention \cite{locatello_object-centric_2020} map inputs into distinct object slots. These structured representations yield compact latent spaces, better generalization, and faster inference, especially in cluttered or unseen environments \cite{chapin_object-centric_2025}. In this work, we adopt a Slot-Attention-based approach and systematically compare it with image-based, PRESTO, and occupancy grid representations (Sec.~\ref{subsec:ablations}).
Building on these principles, our approach integrates object-centric, scene-conditioned diffusion priors with a full-dynamics MPC framework. Unlike prior work, we demonstrate real-time, closed-loop control of a physical robot in cluttered, dynamic environments while strictly enforcing collision and dynamic constraints, highlighting the practical benefits of combining generative priors with model-based control.

\section{DIFFUSION-GUIDED MPC}
\label{sec:diffusion_guided}

We propose a two-stage framework to generate collision-free and dynamically feasible trajectories in real time, even in cluttered and dynamic environments (Fig.~\ref{fig:overview}). The first stage uses a conditional diffusion model to produce a high-quality warm-start trajectory, informed by the current scene and task. This warm-start serves as an effective initialization for the second stage, where a receding-horizon MPC solver refines the trajectory into a locally optimal, dynamically feasible control sequence while enforcing hard collision constraints. By combining learned generative priors with model-based control, our approach leverages the strengths of both: global reasoning from the diffusion model and physical constraint satisfaction from MPC.

\begin{figure}[t]
    \includegraphics[width=1\linewidth]{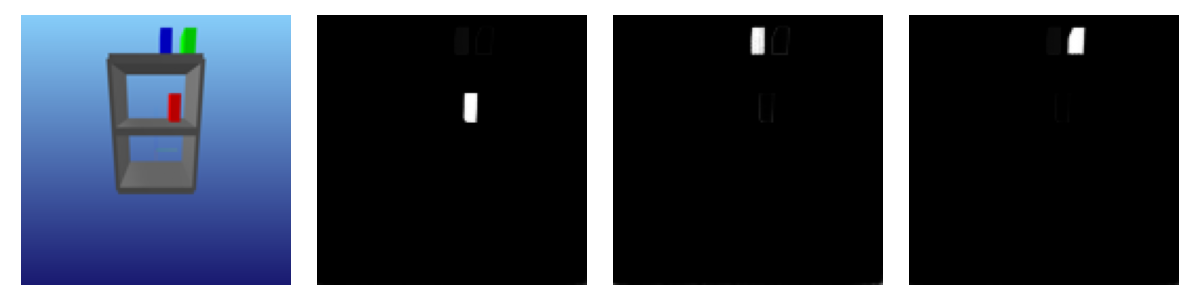}
    \centering
    \caption{
        \textbf{Object-centric scene representation using Slot Attention.} 
        Given an RGB image of the scene, a pre-trained Slot-Attention encoder~\cite{locatello_object-centric_2020} extracts a set of object-centric latent embeddings, called slots, each corresponding to a distinct object or obstacle in the environment. 
        These slots are then used to construct a structured representation of the scene, which serves as input to the diffusion-based trajectory generator. 
        This conditioning enables the model to generate trajectories that account for obstacle geometry and layout, while maintaining generalization to unseen environments.
    }
    \label{fig:slot_attention_masks}
\end{figure}
\subsection{Conditional trajectory generation via diffusion.}
\subsubsection{Trajectory diffusion}

To generate high-quality warm-start trajectories, we train a conditional diffusion model as a learned prior over feasible motion sequences \cite{carvalho_motion_2023, seo_presto_2025}. Our model builds on the Denoising Diffusion Probabilistic Model (DDPM) framework \cite{ho_denoising_2020}, which captures complex, multi-modal distributions through a forward noising and reverse denoising process. A robot trajectory $\traj \!=\! (\conf_0,\dots,\conf_T)$ is treated as a sample from an expert distribution. A forward (noising) process gradually adds Gaussian noise to the trajectory over $K$ steps according to a fixed variance schedule $\{\beta_i\}_{i=1}^K$. The noisy trajectory at step $i$ is given by:

\begin{align}
  \noisytraj_i \mid \noisytraj_0 \sim \mathcal N\!\!\left(\sqrt{\bar\alpha_i}\,\noisytraj_0, (1-\bar\alpha_i)\,\mathbf I\right), \, \bar\alpha_i = \smash{\prod_{j=1}^{i}} (1-\beta_j).
\end{align}
The core of the DDPM is a neural network, $\bm\epsilon_\theta$, trained to predict the noise added at each step. This network parameterizes a reverse (denoising) process that can iteratively construct a clean trajectory sample $\noisytraj$ starting from pure Gaussian noise $\noisytraj_K \sim \mathcal{N}(0, I)$. The network is trained by minimizing a noise-prediction objective:
\begin{equation}
  \mathcal L_{\mathrm{DDPM}} = \mathbb E_{i,\noisytraj_0,\bm\epsilon} \bigl\|\,\bm\epsilon - \bm\epsilon_\theta(\noisytraj_i,i,\mathbf c)\bigr\|_2^2,
  \label{eq:DDPM_loss}
\end{equation}
where $\bm\epsilon$ is the sampled noise and $\mathbf c$ represents the conditioning variables.

\subsubsection{Model Architecture}
We use a Diffusion Transformer (DiT) that denoises a trajectory $\noisytraj_i$ conditioned on the task context. The context includes the robot’s initial configuration $\conf_0$, the Cartesian goal $\goal$, and a scene encoding. The latter is provided by a Slot Attention encoder~\cite{locatello_object-centric_2020}, which converts an RGB image of the workspace into object-centric latent embeddings (slots). This representation captures obstacle information in a compact form, enabling the DiT to generate trajectory proposals that are tailored to each problem instance. We describe the encoder in more detail below.  

\subsubsection{Slot Attention-Based Scene Encoder}
The Slot Attention encoder~\cite{locatello_object-centric_2020} extracts object-centric latent representations from single-view RGB images of the robot workspace. This structured decomposition of the scene into obstacles provides a compact conditioning signal for the diffusion model, enabling generalization across cluttered layouts without relying on fixed spatial encodings or manual annotation.

The encoder consists of a CNN backbone that processes a $128\times128$ RGB image into spatial features, followed by a positional embedding layer and a Slot Attention module with iterative soft clustering. The output is a set of $N$ slots of dimension $D=64$ (typically $N=6$, with 5 foreground objects and 1 background), as illustrated in Fig.~\ref{fig:slot_attention_masks}. Slots are unordered, with each competing to represent part of the scene, yielding a permutation-invariant latent description of obstacles. This permutation invariance of slots prevents the model from relying on fixed object order and thereby improves generalization to novel scenes with varying numbers and configurations of obstacles.

\subsection{Trajectory refinement with MPC.}
\subsubsection{Warm-start rationale}
The diffusion model thus provides trajectory proposals that are close to feasible and well-suited as warm-starts for optimization. However, these outputs do not strictly enforce physical laws or safety constraints. We will empirically show in the next section that the generated trajectories, $\hat{\traj}$, imitate the training data but may still violate torque limits, velocity bounds, or collision-avoidance. Executing them directly on a robot would therefore be unsafe. To transform these guesses into a valid motion trajectory $\refinedtraj$, we refine them by solving an Optimal Control Problem (OCP).
The goal is to find a trajectory that minimizes a cost function~$\ell$ while respecting system dynamics and constraints. The state and control trajectories are simply represented by their temporal discretizations $\statetraj = (\state_0, \ldots, \state_T)$ and $\controltraj = (\control_0, \ldots, \control_{T-1})$  The general OCP is given by:
\begin{subequations}\label{equ:ocp}
\begin{align}
 \underset{\statetraj , \controltraj }{\min} & \sum^{T-1}_{t = 0} \ell_t (\state_t, \control_t) + \ell_T (\state_T)\\
\text{subject to   }
    \state_{t+1} &= f_t(\state_t,\control_t), \\
    c_t(\state_t,\control_t) &\geq 0.
\end{align}
\end{subequations}
Here, $\state_t = (\conf_t, \velocity_t)$ is the robot state, $\control_t$ is the joint torques, $f_t$ is the discretized robot dynamics model and $c_t$ represents hard constraints. 

\subsubsection{MPC costs and constraints}

The running and terminal costs $\ell_t, \ell_T$ combine three terms:  

a. \underline{Goal tracking.} A quadratic penalty drives the end-effector toward its Cartesian target $\goal \in \mathbb{R}^3$. Let $\eepos(\conf_t) \in \mathbb{R}^3$ denote the end effector position (forward kinematics). 
The goal tracking cost is defined as
\begin{equation}
  \ell_{ee}(\state_t) = 
  \|\, \eepos(\conf_t) - \goal \,\|_{Q_{ee}}^2 \, ,
  \label{eq:pos_cost}
\end{equation}
where $Q_{ee} \succeq 0$ is a user-chosen weight matrix that sets the relative importance of errors along the three Cartesian axes.  

b. \underline{State regularization.} To encourage solutions close to the diffusion-based warm-start, we penalize deviations from the diffusion-generated proposal $\hat{\traj}$. The reference state trajectory is  
\begin{equation}
    \noisystatetraj =
    \begin{bmatrix}
        \hat{\conf}_0 & \cdots & \hat{\conf}_T \\
        \hat{\velocity}_0 & \cdots & \hat{\velocity}_T
    \end{bmatrix}, \quad
    \hat{\velocity}_t = \tfrac{\hat{\conf}_{t+1} - \hat{\conf}_t}{\Delta t}.
\end{equation}
The regularization cost is  
\begin{equation}
    \ell_{\state}(\state_t) = \| \state_t -  \hat{\state}_{0,t} \|_{Q_{\state}}^2,
\end{equation}
with $Q_{\state} \succeq 0$ diagonal. This encourages the optimizer to stay close to the diffusion prior while ensuring dynamic feasibility and collision-avoidance through the OCP constraints.  

c. \underline{Control effort.} A quadratic penalty discourages large torques beyond gravity compensation, promoting compliant behavior. The control cost is 
\begin{equation}
    \ell_{\control}(\control_t)
    \;=\;
    \| \control_t - \control_\text{grav}(\conf_t) \|_{Q_u}^2,
\end{equation}
where $\control_t \in \mathbb{R}^n$ are the commanded joint torques at time step $t$, 
$\control_\text{grav}(\conf_t)$ is the gravity compensation torque computed from the robot dynamics model, 
and $Q_u \succeq 0$ is a diagonal weight matrix.  

The term $\control_\text{grav}(\conf_t)$ represents the torques required to balance the manipulator against gravity at configuration $\conf_t$, without inducing any motion.  

\begin{figure*}[t]
    \centering
    \includegraphics[width=0.80\textwidth]{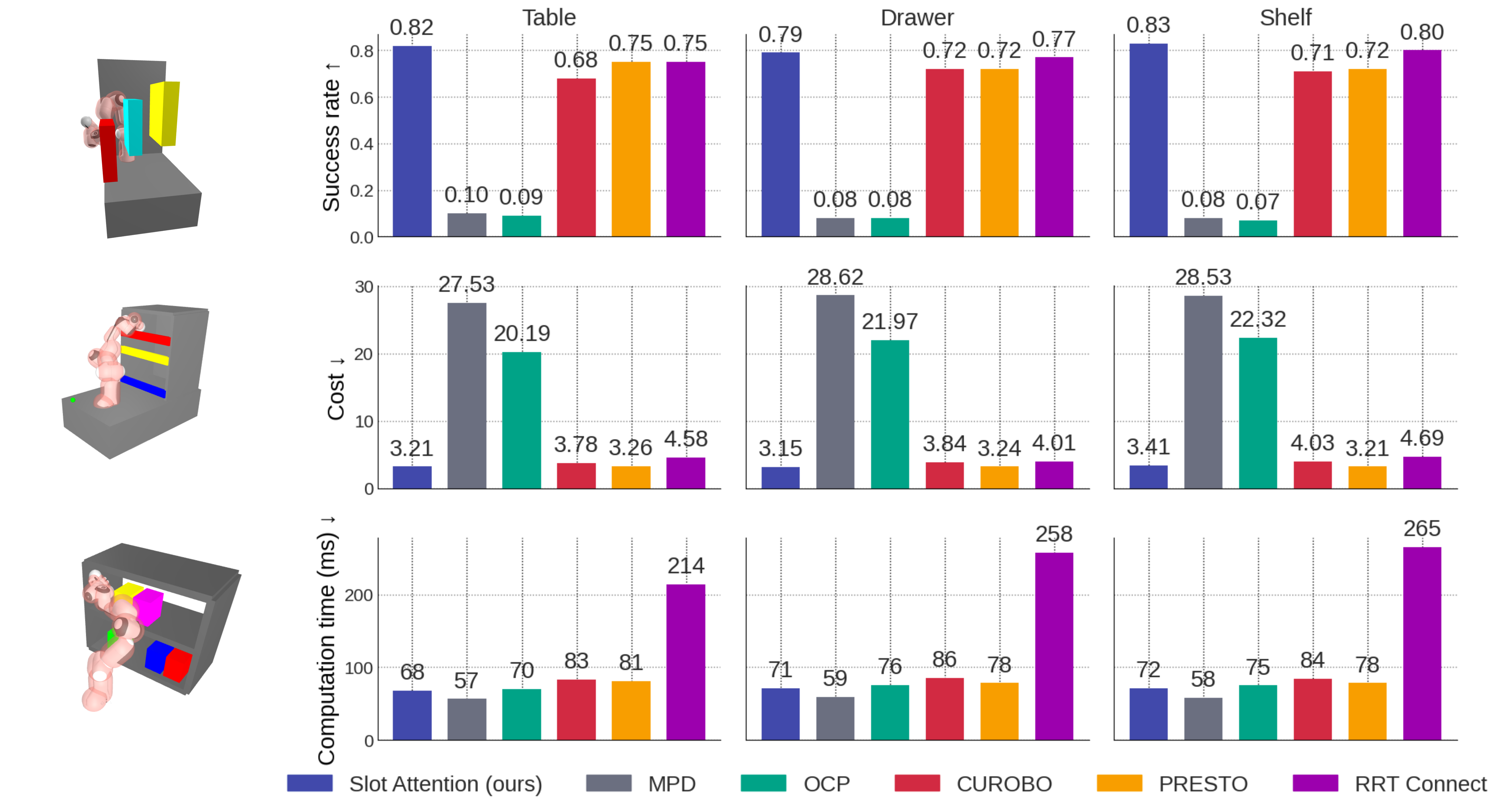}
        \caption{ \textbf{State-of-the-art methods comparison.}  We use Table, Drawer, and Shelf benchmarks shown on the left to compare our diffusion-warm-started MPC against PRESTO~\cite{seo_presto_2025}, Motion-Planning Diffusion (MPD)~\cite{carvalho_motion_2023}, cuRobo \cite{sundaralingam_curobo_2023},  RRT-Connect~\cite{kuffner_rrt-connect_2000} and a DDP-based OCP without a warm-start~\cite{mastalli_crocoddyl_2020}. Metrics are aggregated over 3 different scenes with various obstacles positions. Higher is better for success rate \textbf{(↑)}, lower is better \textbf{(↓)} for average cost and computation time. Our method attains the highest success on all levels, \SI{82}{\percent}, \SI{79}{\percent}, \SI{83}{\percent}, with the lowest costs and sub-\SI{72}{\milli\second} runtime. PRESTO~\cite{seo_presto_2025}, cuRobo~\cite{sundaralingam_curobo_2023} and RRT-Connect~\cite{kuffner_rrt-connect_2000} are slower but still provide high success rates over \SI{70}{\percent}, while MPD and the pure OCP baseline remain below \SI{10}{\percent} success.
  }
    \label{fig:sota_methods_comparison}
\end{figure*}

d. \underline{Collision constraints.} To guarantee safety, we impose a hard collision-avoidance constraint at each timestep following~\cite{haffemayer_model_2024}, requiring the signed-distance $d(\cdot,\cdot)$ between any pair of potentially colliding bodies~$\mathcal B$ to be greater than a safety margin $\epsilon_{\text{safe}}$:
\begin{equation}
    c_{t,i,j}(\state_t) = d\bigl(\mathcal B_i(\state_t),\mathcal B_j(\state_t)\bigr) - \epsilon_{\text{safe}} \ge 0.
    \label{eq:collision_constraint}
\end{equation}

We emphasize that these costs reflect one possible instantiation of our framework, any other task with a standard OCP formulation could be used in place of this choice without altering the overall pipeline.

\subsubsection{Warm-starting the solver}
To avoid convergence to local minima, we use the diffusion model output to warm-start the OCP solver, setting its initial guess to $\noisystatetraj$. For the control trajectory, we initialize the torques using gravity compensation. This directs the solver to a promising region of the search space, accelerating convergence, as we discuss in Sec.~\ref{sec:experiments}.

\subsection{Training data-generation.}
The performance of the generative model is critically dependent on the quality of its training data. To learn a useful motion prior, the model must be trained on a diverse dataset of expert trajectories that are not only collision-free but also dynamically smooth and efficient. Trajectories from purely geometric planners like RRT~\cite{lavalle_rapidly-exploring_1998} are often jerky and not directly executable. Therefore, a dedicated data-generation pipeline is required to produce high-quality examples.
We construct our training dataset by generating thousands of random scenes in  simulated environments described in Sec.~\ref{sec:experiments}. For each scene, we have computed multiple expert trajectories. We first sample a start/goal pair and generate a geometrically valid path using RRT. This path is then used to initialize and solve the OCP defined in~\eqref{equ:ocp}, resulting in a smooth, feasible expert trajectory. In parallel, the Slot Attention algorithm was trained beforehand on a task-specific dataset. The diffusion model is then trained on this dataset by minimizing the loss in~\eqref{eq:DDPM_loss}.

\section{EXPERIMENTS}\label{sec:experiments}
\subsection{Experimental setup.}\label{subsec:sim_setup}
We evaluate our method on three distinct motion-planning tasks involving a Franka Emika Panda robotic arm operating in cluttered environments, populated with varying numbers of obstacles, as illustrated in Fig.~\ref{fig:teaser}.  
These setups are particularly challenging due to the highly non-convex geometry of the obstacles and the cluttered nature of the workspace. We first conduct comparative benchmarks and analyses (Sec.~\ref{subsec:ablations}), and then validate on real hardware (Sec.~\ref{subsec:real_robot}). We define three benchmark environments that differ in geometry and obstacle layout: a table environment, a drawer environment, and a shelf environment, as shown in Fig.~\ref{fig:sota_methods_comparison}.
Each environment is instantiated with 15 distinct obstacle configurations, varying both the number and placement of obstacles, with a maximum of three obstacles in the table and drawer environments and up to five in the shelf environment. For every configuration, we generate 30 planning problems by sampling initial robot configurations and end effector goals, resulting in a broad and diverse set of challenging scenarios.

\subsubsection{Training dataset} 
For training the diffusion process, we generated datasets by randomly sampling scenes from the simulated benchmark environments. For the shelf environment, the dataset contains 300 scenes with 1 to 6 randomly placed box-shaped obstacles. For each scene, we generate 100 random pairs of initial configurations and end effector goals, producing a total of 30 000 trajectories. For the drawer and table environments, we generate 100 scenes each, with 1 to 3 randomly placed obstacles per scene and 100 planning problems per scene, resulting in 10 000 trajectories for each environment. All trajectories are obtained using RRT and subsequently refined using an OCP solver. Each trajectory contains a fixed number of nodes, with $T = 50$.

\subsubsection{Metrics and protocol}
To evaluate the quality of the generated trajectories, we define three metrics. First, the success rate measures the percentage of trajectories that reach the goal without collisions, where higher values indicate better performance.
Second, the penetration depth quantifies the cumulative interpenetration between the robot and obstacles along the trajectory; lower values correspond to improved collision-avoidance, and a successful trajectory yields zero penetration.  
Finally, the average cost corresponds to the total OCP cost, reflecting both feasibility and proximity to the goal.
For each method, we sample a batch of 10 trajectories per scene and report the one with the best overall performance according to the evaluation metrics. The benchmarks were done on an Intel Core i9-14900K, 64 GB RAM, and an NVIDIA RTX 5000 Ada GPU with 32 GB VRAM.

\subsection{State-of-the-art comparison.}
We evaluate our approach against a range of classical and learning-based motion-planning methods to demonstrate that combining diffusion-based warm-starting with OCP refinement yields superior computational efficiency and higher-quality trajectories. The first learning-based baseline is PRESTO~\cite{seo_presto_2025}, a diffusion model operating in configuration space, where the scene is represented by sampling a predefined set of robot configurations. For each configuration, a binary collision indicator is computed, and the resulting indicator vector is concatenated to form the conditioning input. The second baseline is Motion-Planning Diffusion (MPD) \cite{carvalho_motion_2023}, a learning-based approach where an unconditional diffusion model learns to generate motion trajectories, with collision-avoidance encouraged through a cost gradient-based guidance within the diffusion process. For fairness, we adapt both of their architecture and training procedure to match our dataset and task setup.  
We then compare our approach to both classical and optimization-based planners. RRT Connect~\cite{kuffner_rrt-connect_2000} is a widely-used bidirectional sampling-based planner, implemented in Python with C++ bindings, that incrementally builds trees from the start and goal configurations and attempts to connect them. CuRobo~\cite{sundaralingam_curobo_2023} is a recent GPU-accelerated framework that combines trajectory optimization with sampling strategies to efficiently handle high-dimensional planning problems. To ensure a fair comparison under the same time budget, we terminate all algorithms after a fixed number of iterations.

Finally, to demonstrate the benefit of warm-starting the OCP, we compare our method to a standard OCP solver~\cite{mastalli_crocoddyl_2020}, which solves the full optimal control problem defined in~\eqref{equ:ocp} using Differential Dynamic Programming (DDP)~\cite{mayne_differential_1973, tassa_synthesis_2012} without any warm-start. This method directly optimizes dynamic feasibility and task objectives but remains highly sensitive to local minima.  

Fig.~\ref{fig:sota_methods_comparison} compares our method to the baselines in terms of success rate, average cost, and computation time. Across all scenes, our approach achieves the highest success rates (\SI{82}{\percent}, \SI{79}{\percent}, and \SI{83}{\percent}), slightly outperforming PRESTO, cuRobo, and RRT-Connect, and clearly surpassing MPD and the pure OCP baseline which remain below 10 percent. The low performance of OCP and MPD is expected, since highly non-convex and cluttered scenes often prevent the OCP from converging to feasible solutions, and MPD fails to generalize across varying obstacle layouts.

\begin{figure}[t]
    \centering
    \includegraphics[width=0.45\textwidth]{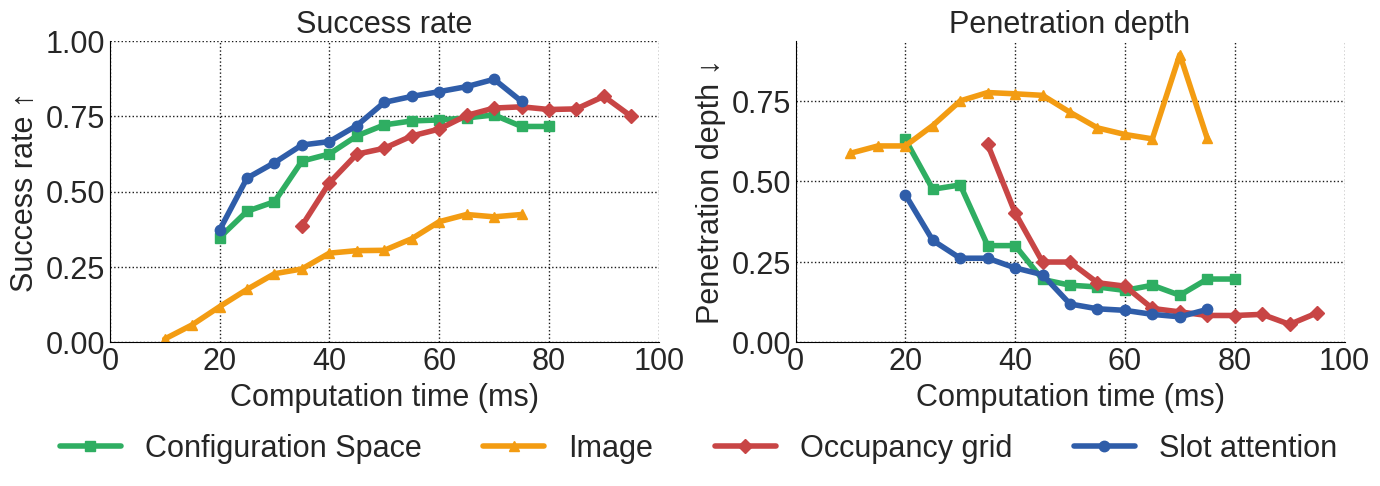}
    \caption{\textbf{Ablation of scene representations.}
 We evaluated our control architecture using different scene representations. The Slot Attention-based representation achieved the best performance, especially in low-computation scenarios. The configuration-space representation performed nearly as well, while image-based conditioning was uncompetitive for control applications.
    }
    \label{fig:conditioning_methods_comparison}
\end{figure}
Our method also achieves the lowest trajectory costs, producing smoother and more dynamically consistent motions. Runtime stays below \SI{72}{\milli\second}, which is slightly faster than the other baselines, while RRT-Connect requires more time and is not suitable for real time settings.

Overall, combining diffusion based planning with OCP refinement and a slot based latent obstacle representation provides the best balance of success, safety, and efficiency. The latent encoding gives the model a compact scene representation, which keeps the architecture lightweight and enables fast inference while still capturing the relevant geometry. The diffusion model then produces a full trajectory that is consistent with this latent scene encoding, but its samples can still contain local inaccuracies or near collision segments. The OCP refinement corrects these issues by solving a joint optimization over dynamics, limits, and collision constraints, allowing it to adjust entire trajectory segments consistently rather than applying local reactive corrections, as simpler low level controllers do. This produces smooth and feasible motions with high success rates while keeping computation time low.

As a result, our method consistently outperforms both sampling based and purely model based approaches in cluttered environments.
\begin{figure}[t]
    \centering
    \includegraphics[width=0.45\textwidth]{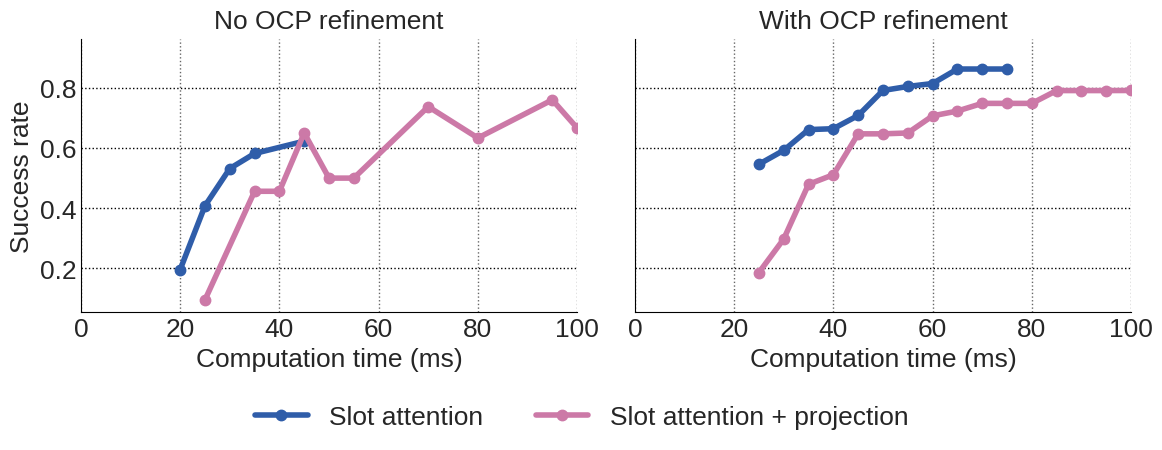}
    \caption{\textbf{Impact of the cost gradient-based guidance.} Comparing the impact of the cost gradient-based guidance at each diffusion step, as used in \cite{carvalho_motion_2023}.
    Results show that guidance is not helpful with a low computational budget, given by control requirements.}
    \label{fig:ablation_projection}
\end{figure}

\begin{figure}[t]
    \centering
    \begin{subfigure}{0.23\textwidth}
        \centering
        \includegraphics[width=\linewidth]{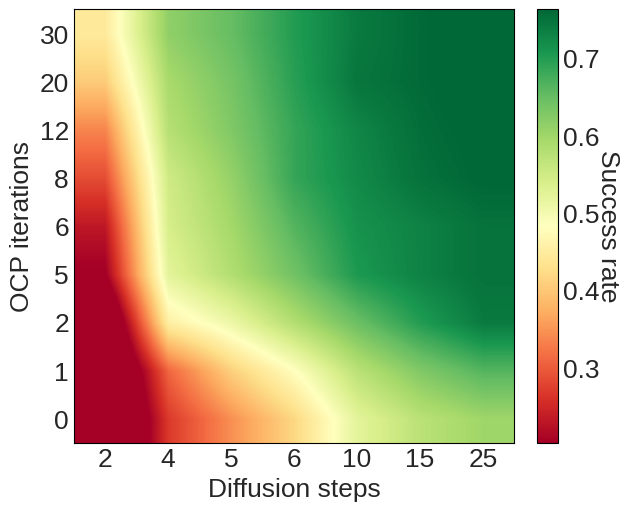}
        \caption{Success rate}
        \label{fig:heatmaps:success}
    \end{subfigure}
    \begin{subfigure}{0.23\textwidth}
        \centering
        \includegraphics[width=\linewidth]{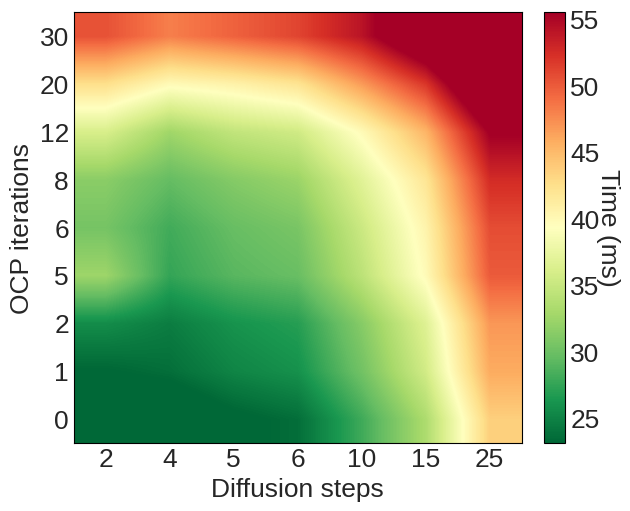}
        \caption{Computation time (ms)}
        \label{fig:heatmaps:time}
    \end{subfigure}
    \caption{\textbf{Success rate and computation time vs. diffusion steps and OCP iterations.} 
    Performance improves with more diffusion refinement and OCP iterations (greener indicates better outcomes).}
    \label{fig:heatmaps}
    \vspace{-2mm}
\end{figure}

\subsection{Ablation studies.}\label{subsec:ablations}
\subsubsection{Latent scene representation comparison}
We evaluate several scene encoding strategies to assess their impact on downstream trajectory generation. We first describe the conditioning methods taken from the literature, and then compare them across scene encoding strategies. The first approach relies on an occupancy-grid representation, where the 3D workspace is discretized into a voxel grid, and each cell is labeled as free (0) or occupied (1). This binary occupancy map is then encoded using a 3D CNN with frozen weights. The second approach adopts an image-based representation, where a single RGB image of the scene, captured from a fixed viewpoint, is processed by a ResNet-18 encoder \cite{he_deep_2016} to produce a compact scene embedding. The third approach is the configuration-space representation from PRESTO~\cite{seo_presto_2025}. Finally, we introduce our method, referred to as Slot Attention conditioning, which leverages an object-centric latent representation obtained through unsupervised segmentation using Slot Attention. The resulting slots are directly used as conditioning inputs to the diffusion model. Fig.~\ref{fig:conditioning_methods_comparison} compares these four conditioning strategies.
Our approach achieves the highest success rates, especially in low- to mid-budget regimes (\SI{20}{\milli\second}-\SI{60}{\milli\second}), which is mostly relevant for online control. For example, we reach \SI{80}{\percent} success within \SI{50}{\milli\second}, whereas occupancy grids and the configuration space require over \SI{70}{\milli\second} to reach \SI{75}{\percent} of success. Image-based conditioning performs worst overall, saturating below \SI{50}{\percent}, highlighting the importance of structured, object-centric scene representations. Our method also achieves the lowest collision rates and shallowest penetration depths at most budgets, confirming its advantage in generating safer trajectories.

\subsubsection{Impact of the cost gradient-based guidance}
Fig.~\ref{fig:ablation_projection} evaluates the impact of adding a cost gradient-based guidance  toward the feasible manifold at each denoising step, following \cite{carvalho_motion_2023}. Without OCP refinement, the cost gradient-based guidance improves success at low budgets by steering trajectories closer to feasibility early on. However, when OCP refinement is applied, its benefit diminishes and even harms performance at higher budgets due to unnecessary overhead. We thus omit the cost gradient-based guidance in our final pipeline.

\subsubsection{Trade-off between computation and warm-start quality} Fig.~\ref{fig:heatmaps} analyzes the effect of diffusion steps and OCP iterations on success rate and computation time.
Success improves rapidly when increasing diffusion steps from 3 to 8 and OCP iterations up to 10, reaching \SI{60}{\percent}-\SI{70}{\percent}. Beyond 10 steps and 15 iterations, success rates plateau around \SI{80}{\percent}, showing diminishing returns despite higher runtimes. Computation time grows almost linearly with diffusion steps, while OCP iterations contribute modestly until exceeding 10, after which runtimes increase steeply. The optimal combination of diffusion steps and OCP iterations therefore depends on the available computational budget and the specific task requirements. 
Fig.~\ref{fig:heatmaps} can thus guide the selection of the most suitable trade-off.

\begin{figure}[t]
  \centering
  \captionsetup[subfigure]{aboveskip=2pt,belowskip=2pt,justification=centering}
  
  \begin{subfigure}{0.15\textwidth}
    \includegraphics[width=\linewidth]{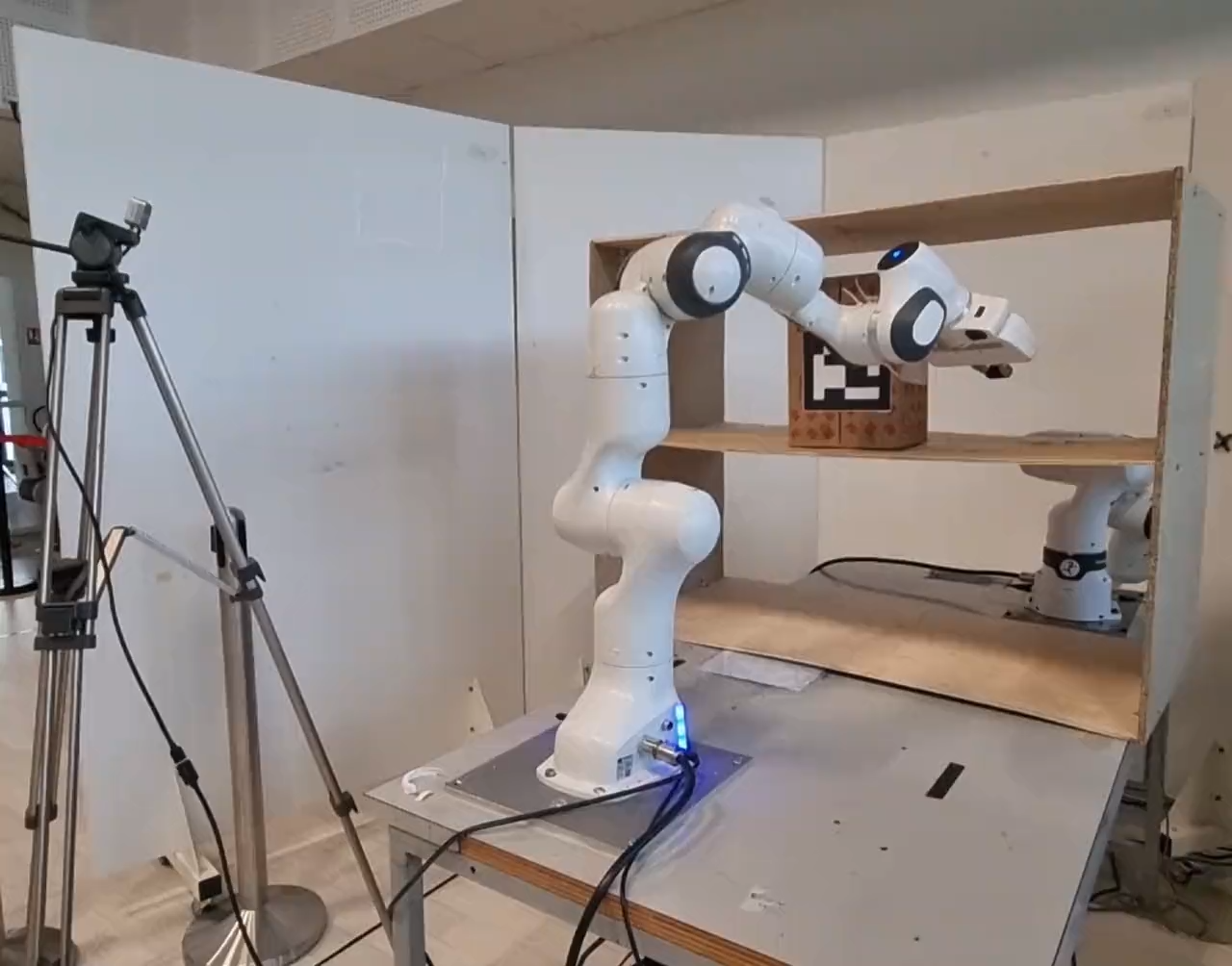}
  \end{subfigure}
  \begin{subfigure}{0.15\textwidth}
    \includegraphics[width=\linewidth]{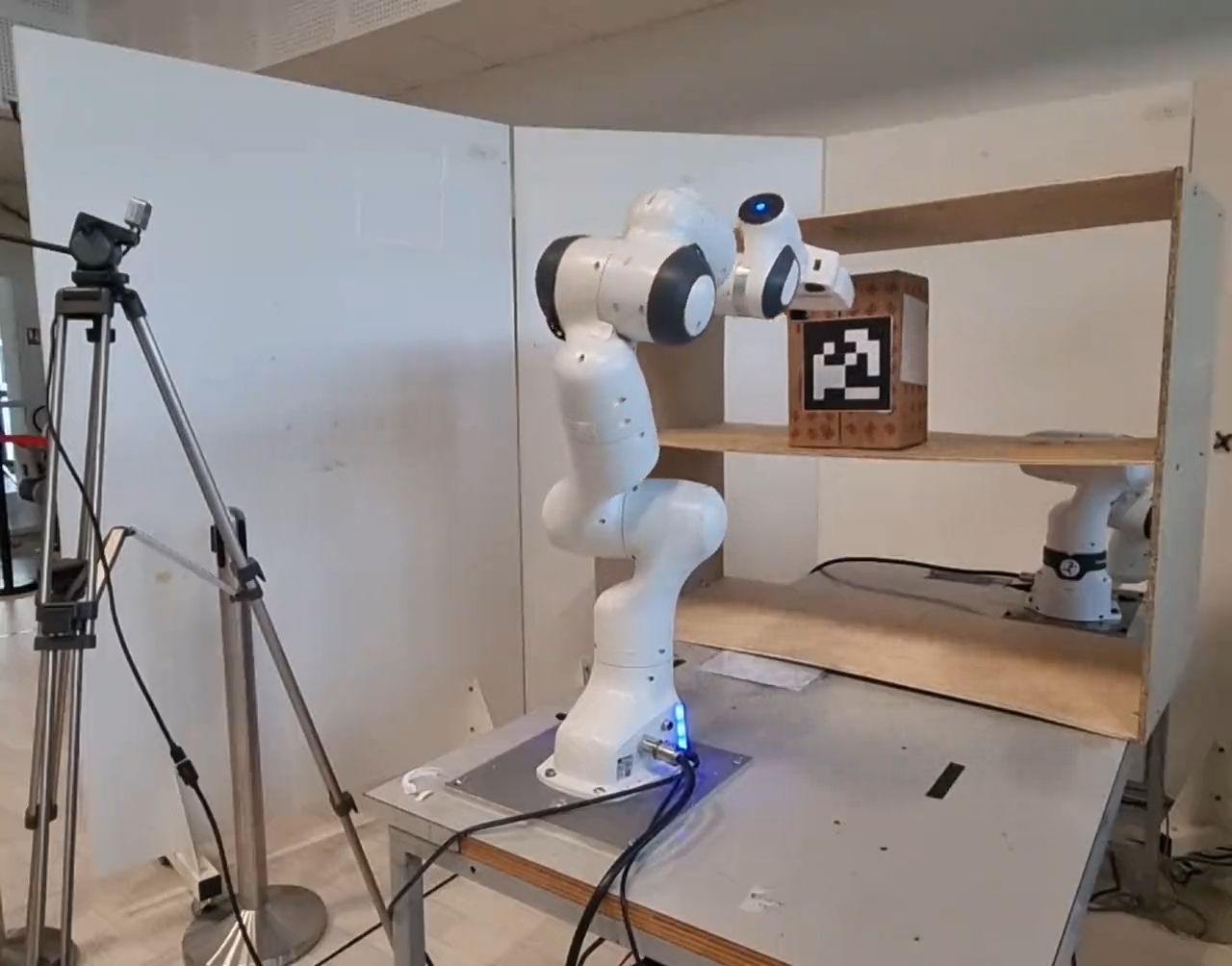}
  \end{subfigure}
  \begin{subfigure}{0.15\textwidth}
    \includegraphics[width=\linewidth]{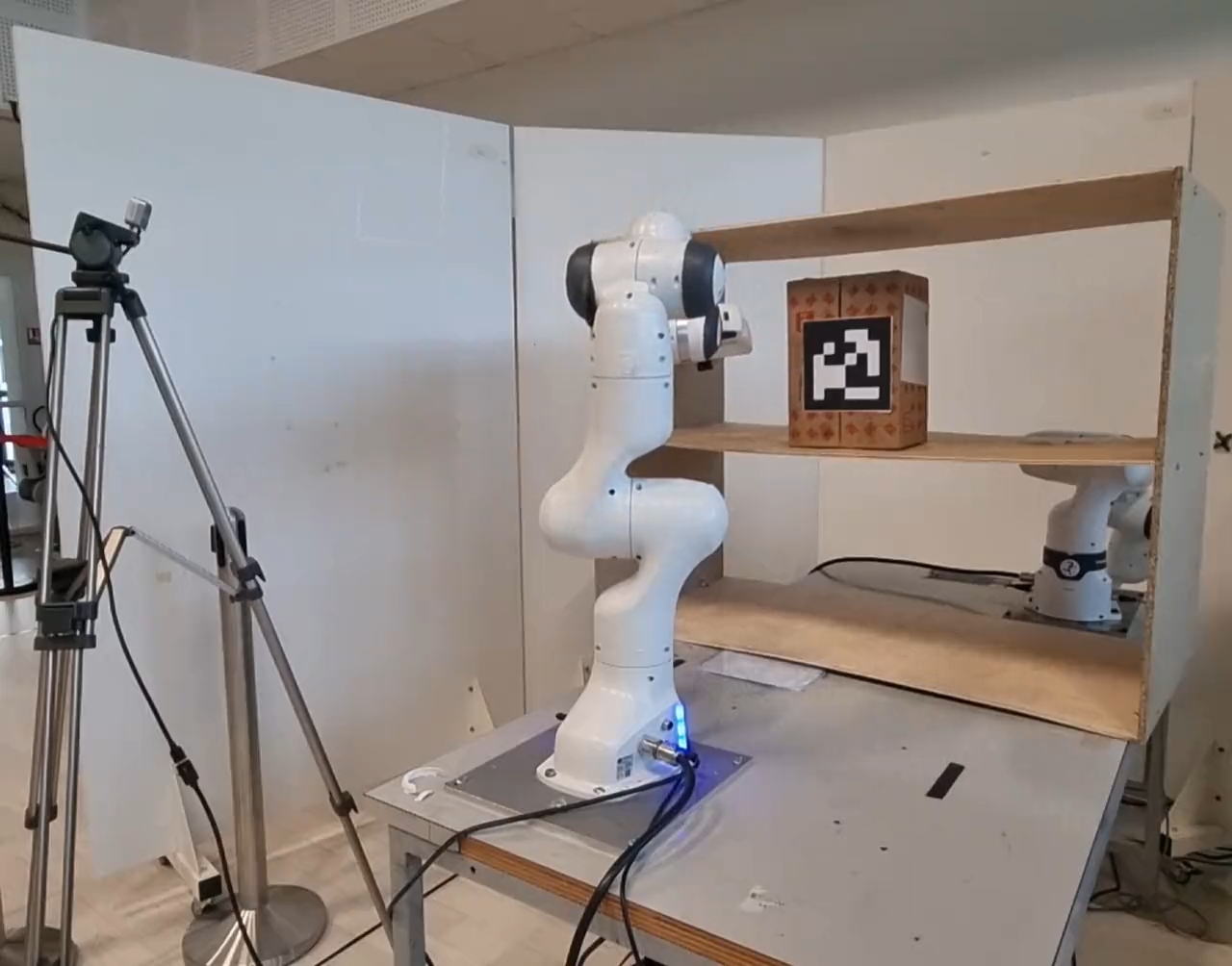}
  \end{subfigure}

  \hfill

  \begin{subfigure}{0.15\textwidth}
    \includegraphics[width=\linewidth]{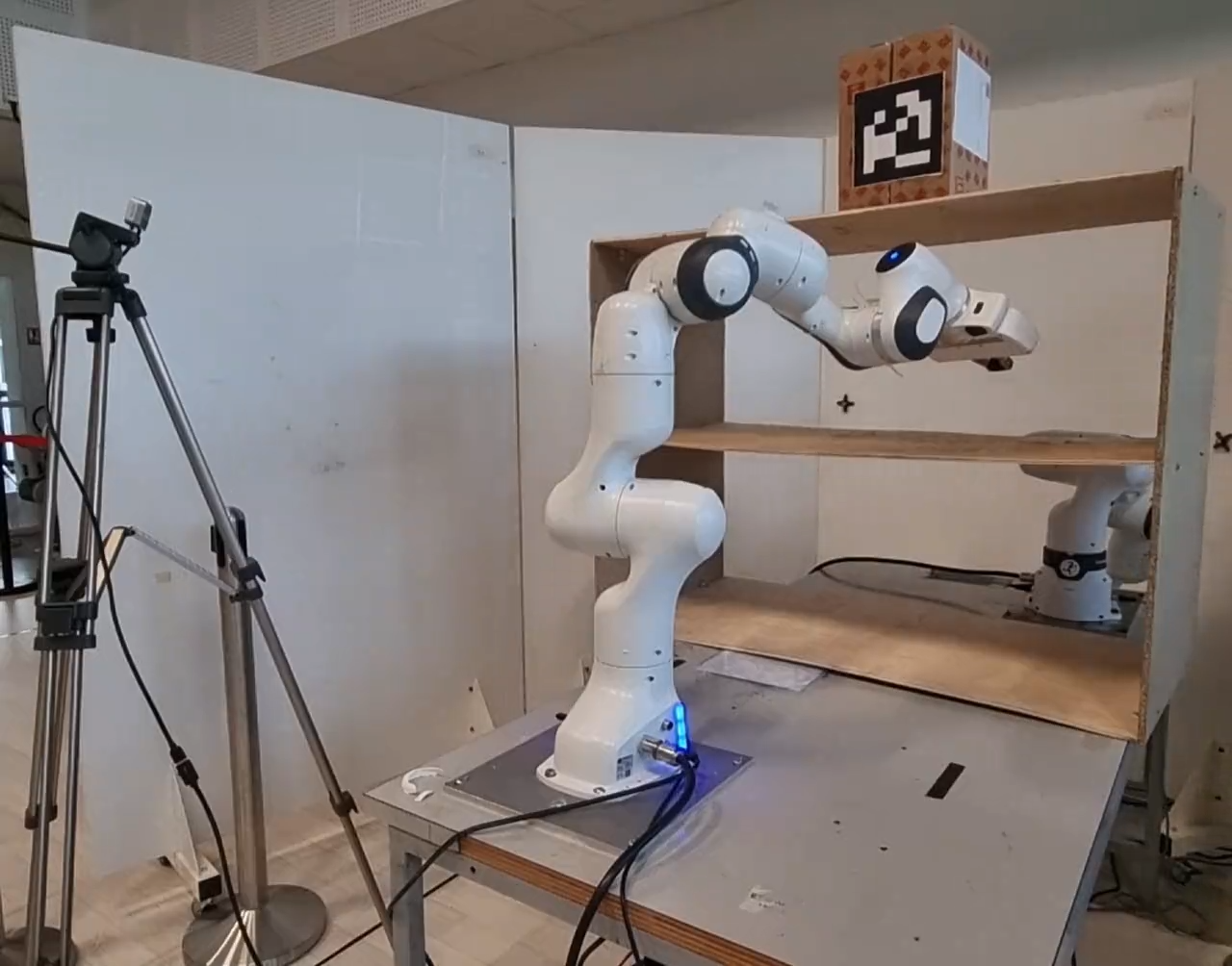}
  \end{subfigure}
  \begin{subfigure}{0.15\textwidth}
    \includegraphics[width=\linewidth]{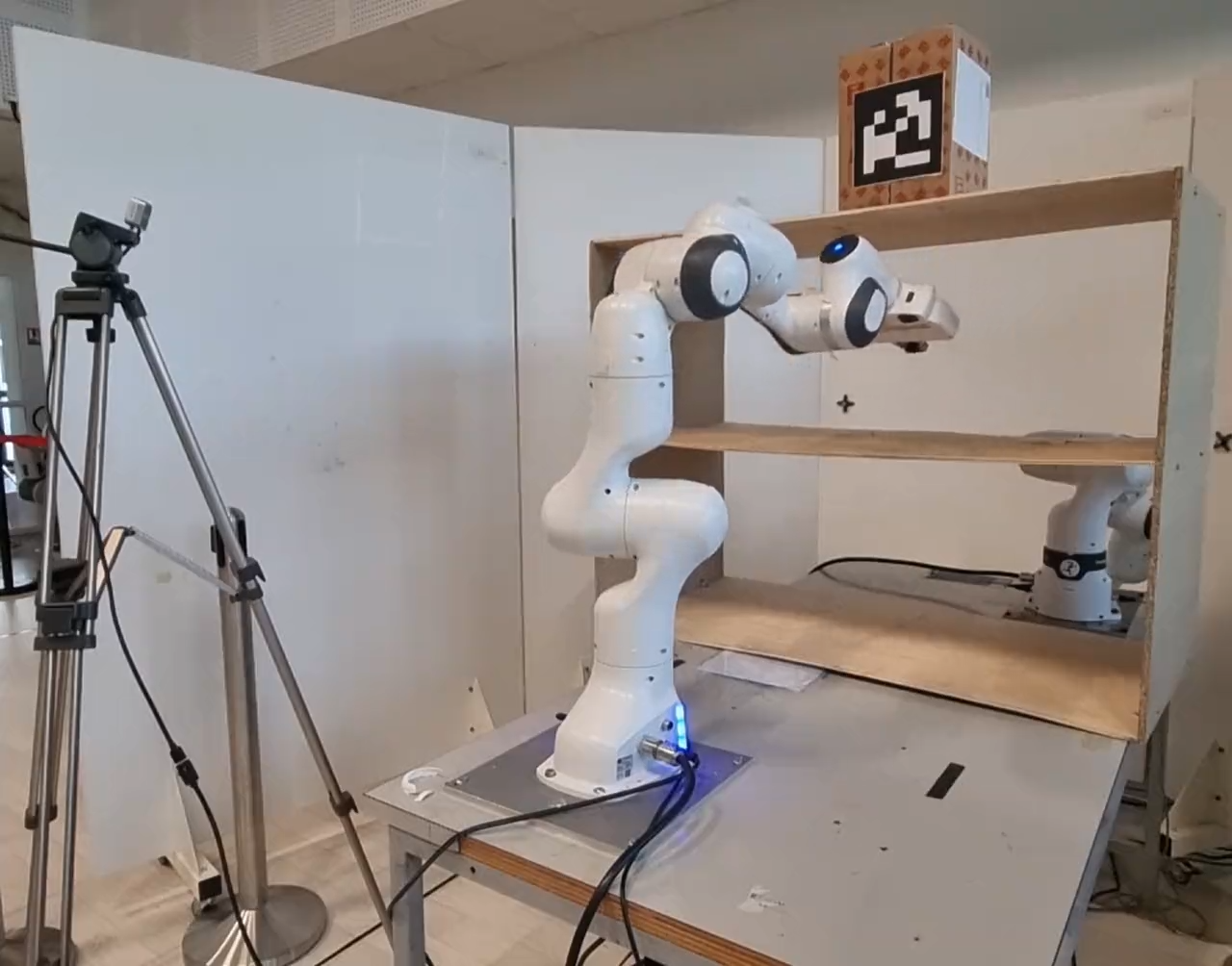}
  \end{subfigure}
  \begin{subfigure}{0.15\textwidth}
    \includegraphics[width=\linewidth]{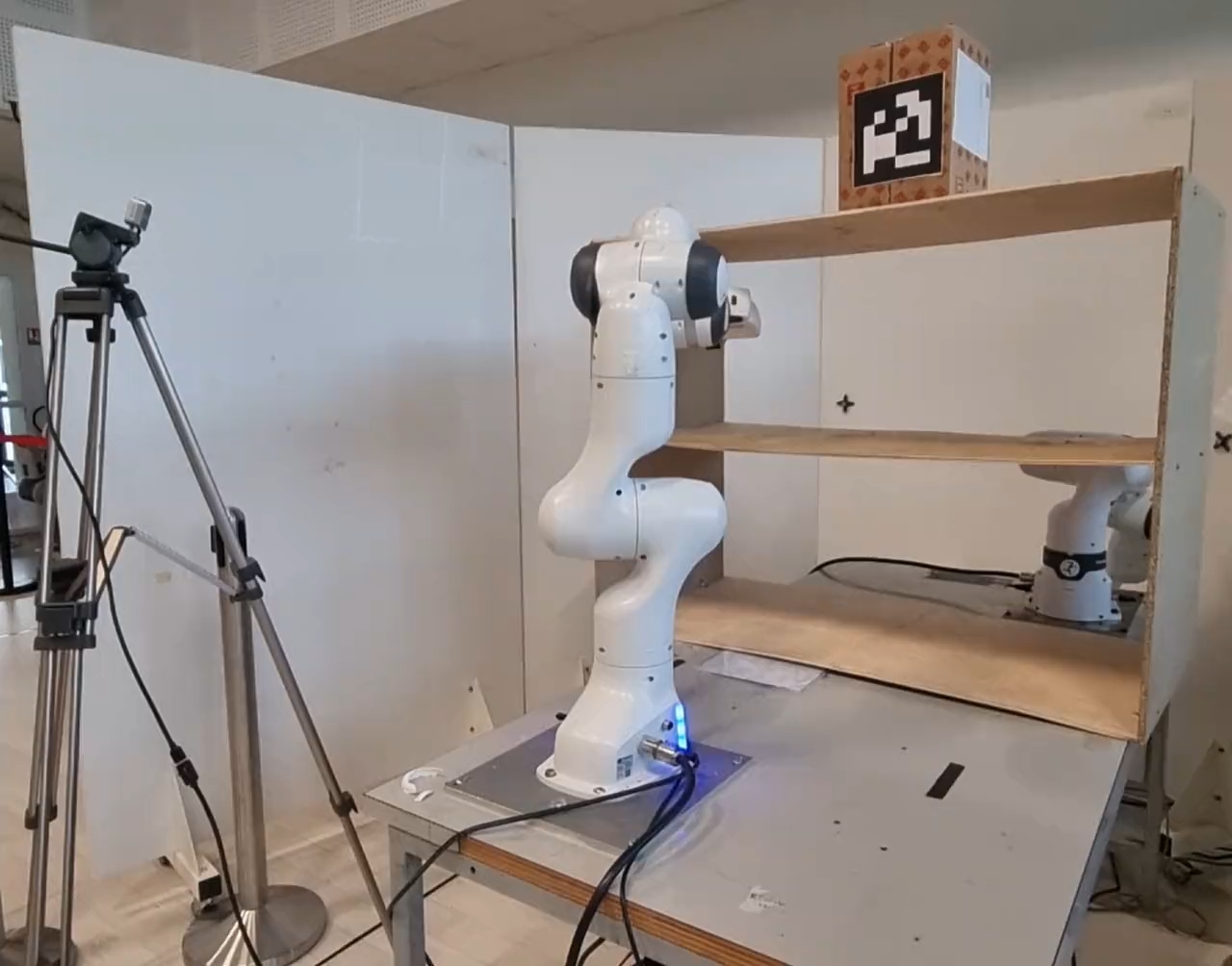}
  \end{subfigure}
  
  \caption{
    \textbf{Real-robot experiments.} 
The robot is tasked with moving from left to right within the middle shelf. As shown in the top row, our approach successfully avoids an obstacle detected via AprilTags while completing the task. When the obstacle is removed on the fly (bottom row), the robot seamlessly re-optimizes its motion, reaching the target more efficiently.
  }
  \label{fig:real_robot_experiments}
\end{figure}

\subsection{Experimental validation on the real robot.}\label{subsec:real_robot}
Experiments were conducted on a Franka Emika Panda robotic arm controlled via torque commands. 
The MPC framework was implemented in Python with C++ bindings and executed on an AMD Ryzen~9~5950X CPU at \SI{3.4}{\giga\hertz}. 
The dynamics of the system were computed using \texttt{pinocchio} \cite{carpentier_pinocchio_2019}, and in our setup the dynamics are time-invariant. The optimal control problem (OCP) was solved using \texttt{mim-solvers}~\cite{jordana_stagewise_2025} inside \texttt{crocoddyl}~\cite{mastalli_crocoddyl_2020}, consistent with the simulation setup described in Section~\ref{subsec:sim_setup}. 
The robot dynamics were modeled using inertial parameters from~\cite{gaz_dynamic_2019}. Obstacle detection and pose estimation were performed in real-time using AprilTags~\cite{olson_apriltag_2011}, with known obstacle dimensions and external-camera-based object tracking. Control commands were computed at \SI{1}{\kilo\hertz}, while the OCP was solved at \SI{100}{\hertz} with a prediction horizon of \(N_h = 15\) nodes and a time step of \({\Delta t} = \SI{10}{\milli\second}\), resulting in an optimization horizon of \SI{150}{\milli\second}. 
At each OCP iteration, we compute the optimal trajectory and the corresponding Riccati gains \cite{dantec_first_2022}. The Riccati gains are used to apply a low-level feedback policy at \SI{1}{\kilo\hertz} until the next replanning time. 
A collision safety margin of \SI{1}{\centi\meter} was enforced via a lower-bound constraint on the signed-distance. The MPC operated under a receding-horizon scheme. 
Whenever a new high-level task was issued (e.g., reaching a new target position), the diffusion model generated a warm-start trajectory, replacing the previous solution. 
As Fig.~\ref{fig:real_robot_experiments} shows, this warm-started OCP enabled reactive and efficient collision-avoidance in cluttered and dynamically changing environments.

\section{CONCLUSION}
We presented a diffusion-based framework for generating fast, high-quality trajectory proposals that warm-start a model-based MPC solver, improving convergence while maintaining low-latency inference. The proposed approach was validated on a Franka Emika Panda robot in cluttered shelf environments, and the open-source code along with a video demonstration of the real-time system are provided in the \href{https://ahaffemayer.github.io/diffusion_warmstart_slot/}{supplementary materials}.

While our current conditioning relies on Slot Attention trained on relatively flat, 2D scenes, the framework is designed to naturally extend to fully 3D settings. Future work will explore richer object-centric representations to further improve generalization. Another exciting direction is an online variant that integrates diffusion directly within the MPC loop and adaptively selects between the previous plan and the newly inferred one, enabling even faster and more responsive control and providing a way to enforce safety guarantees on diffusion policies, which can be integrated with large vision–language action models such as $\pi_{0.5}$\cite{intelligence_05_2025}. We believe these extensions will significantly broaden the applicability of our approach to more complex, dynamic, and cluttered environments.





\bibliographystyle{IEEEtran}
\bibliography{IEEEabrv, RAL2025}
\end{document}